\documentclass[11pt,a4paper]{article}
\usepackage[hyperref]{acl2021}
\usepackage{times}
\usepackage{latexsym}

\usepackage{microtype}

\aclfinalcopy 

\usepackage{CJKutf8} 

\usepackage{times}
\usepackage{epsfig}
\usepackage{graphicx}
\usepackage{amsmath}
\usepackage{amssymb}
\usepackage{array}

\usepackage{longtable}

\definecolor{demphcolor}{RGB}{144,144,144}

\makeatletter
\newcommand*{\@rowstyle}{}

\newcommand*{\rowstyle}[1]{
  \gdef\@rowstyle{#1}%
  \@rowstyle\ignorespaces%
}

\newcolumntype{=}{
  >{\gdef\@rowstyle{}}%
}

\newcolumntype{+}{
  >{\@rowstyle}%
}
\makeatother

\makeatletter
\def\thickhline{%
  \noalign{\ifnum0=`}\fi\hrule \@height \thickarrayrulewidth \futurelet
   \reserved@a\@xthickhline}
\def\@xthickhline{\ifx\reserved@a\thickhline
               \vskip\doublerulesep
               \vskip-\thickarrayrulewidth
             \fi
      \ifnum0=`{\fi}}
\makeatother

\newlength{\thickarrayrulewidth}
\usepackage{multirow}
\usepackage{booktabs}

\usepackage{pifont}
\newcommand{\xdownarrow}[1]{%
  {\left\downarrow\vbox to #1{}\right.\kern-\nulldelimiterspace}
}

\usepackage{xcolor}
\definecolor{ForestGreen}{rgb}{0.13, 0.55, 0.13}
\definecolor{bittersweet}{rgb}{1.0, 0.44, 0.37}
\definecolor{orchid}{rgb}{0.478, 0.506, 1.0}
\definecolor{salmon}{rgb}{0.968, 0.503, 0.408}
\definecolor{green}{rgb}{0.438, 0.678, 0.408}
\definecolor{orange}{rgb}{1.0, 0.75, 0.}

\usepackage{colortbl}
\definecolor{Gray}{gray}{0.85}

\def\DsetName{\textsc{mTVR}}
\def\ModelName{mXML}

\title{\DsetName: Multilingual Moment Retrieval in Videos}

\author{
  Jie Lei $\;\;\;\;\;$ 
  Tamara L. Berg $\;\;\;\;\;$ Mohit Bansal \\
  Department of Computer Science \\ University of North Carolina at Chapel Hill \\
  {\tt \{jielei, tlberg, mbansal\}@cs.unc.edu} \\
}

\begin{document}
\maketitle
\begin{abstract}
We introduce \DsetName, a large-scale multilingual video moment retrieval dataset, containing 218K English and Chinese queries from 21.8K TV show video clips.
The dataset is collected by extending the popular TVR dataset (in English) with paired Chinese queries and subtitles. 
Compared to existing moment retrieval datasets, \DsetName~is multilingual, larger, and comes with diverse annotations.
We further propose \ModelName, a multilingual moment retrieval model that learns and operates on data from both languages, via encoder parameter sharing and language neighborhood constraints.
We demonstrate the effectiveness of \ModelName~on the newly collected \DsetName~ dataset, where \ModelName~outperforms strong monolingual baselines while using fewer parameters.
In addition, we also provide detailed dataset analyses and model ablations.
Data and code are publicly available at \url{https://github.com/jayleicn/mTVRetrieval}
\end{abstract}

\section{Introduction}\label{introuction}
\begin{figure}[!t]
\begin{center}
  \includegraphics[width=0.99\columnwidth]{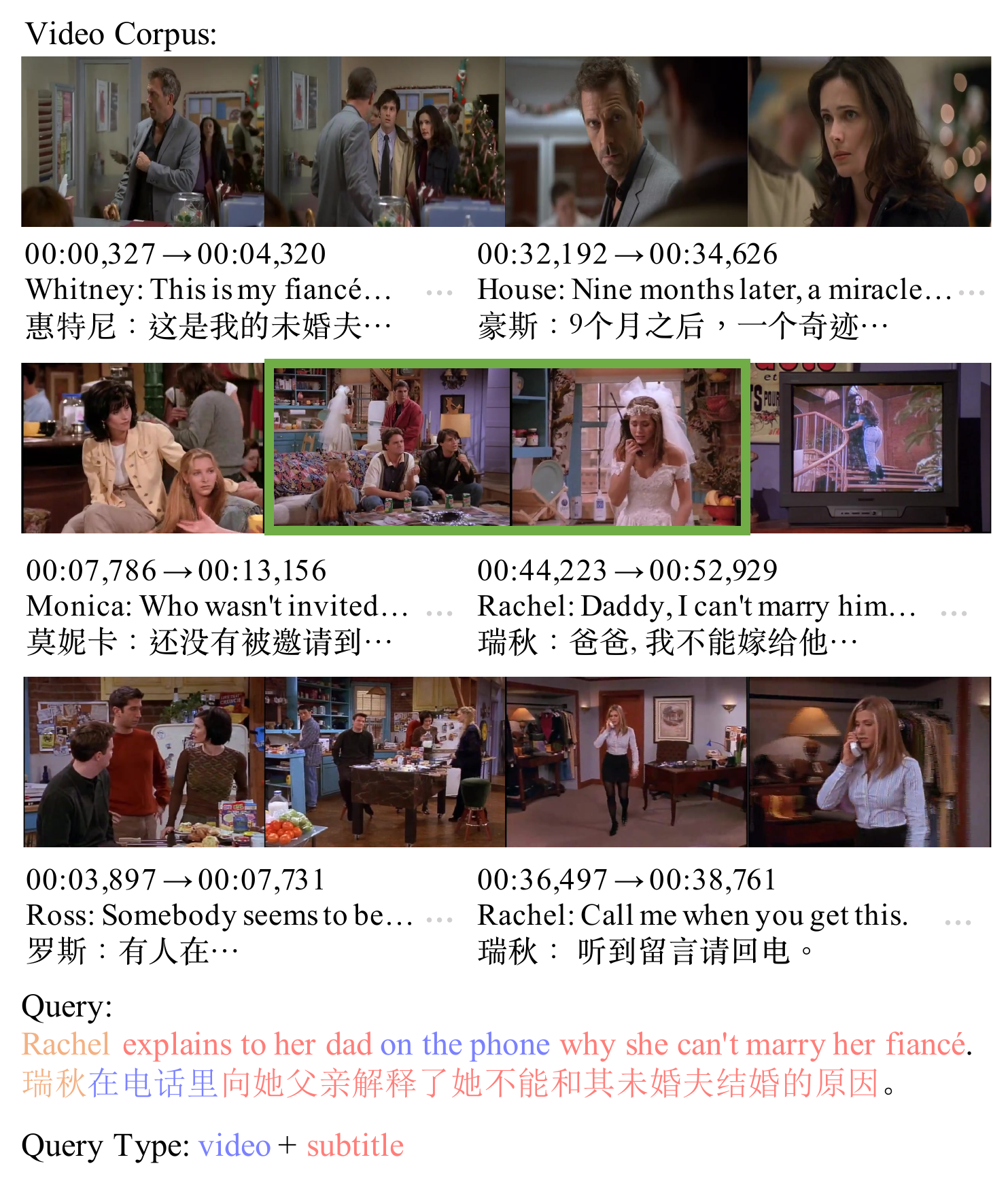}
  \vspace{-12pt}
  \caption{
  A \DsetName~example in the Video Corpus Moment Retrieval (VCMR) task. Ground truth moment is shown in \textit{\textcolor{green}{green}} box. Colors in the query text indicate whether the words are more related to video (\textcolor{orchid}{orchid}) or  subtitle (\textcolor{salmon}{salmon}) or both (\textcolor{orange}{orange}). 
  The query and the subtitle text are presented in both English and Chinese. 
  The video corpus typically contains thousands of videos, for brevity, we only show 3 videos here.
  }
  \label{fig:data_example}
  \end{center}
\end{figure}

The number of videos available online is growing at an unprecedented speed.
Recent work~\cite{escorcia2019temporal,lei2020tvr} introduced the Video Corpus Moment Retrieval (VCMR) task: given a natural language query, a system needs to retrieve a short moment from a large video corpus. 
Figure~\ref{fig:data_example} shows a VCMR example.
Compared to the standard text-to-video retrieval task~\cite{xu2016msr,yu2018joint}, it allows more fine-grained moment-level retrieval, as it requires the system to not only retrieve the most relevant videos, but also localize the most relevant moments inside these videos. 
Various datasets~\cite{Krishna2017DenseCaptioningEI,anne2017localizing,gao2017tall,lei2020tvr} have been proposed or adapted for the task. 
However, they are all created for a single language (English), though the application could be useful for users speaking other languages as well. 
Besides, it is also unclear whether the progress and findings in one language generalizes to another language~\cite{bender2009linguistically}.
While there are multiple existing multilingual image datasets~\cite{gao2015you,elliott-etal-2016-multi30k,shimizu2018visual,pappas2016multilingual,lan2017fluency,li2019coco}, the availability of multilingual video datasets~\cite{Wang_2019_ICCV,chen2011collecting} is still limited.

Therefore, we introduce~\DsetName, a large-scale, multilingual moment retrieval dataset, with 218K human-annotated natural language queries in two languages, English and Chinese. 
\DsetName~extends the TVR~\cite{lei2020tvr} dataset by collecting paired Chinese queries and Chinese subtitle text (see Figure~\ref{fig:data_example}).
We choose TVR over other moment retrieval datasets~\cite{Krishna2017DenseCaptioningEI,anne2017localizing,gao2017tall} because TVR is the largest moment retrieval dataset, and also has the advantage of having dialogues (in the form of subtitle text) as additional context for retrieval, in contrast to pure video context in the other datasets.
We further propose \ModelName, a compact, multilingual model that learns jointly from both English and Chinese data for moment retrieval. 
Specifically, on top of the state-of-the-art monolingual moment retrieval model XML~\cite{lei2020tvr}, we enforce encoder parameter sharing~\cite{sachan2018parameter,dong2015multi} where the queries and subtitles from the two languages are encoded using shared encoders. 
We also incorporate a language neighborhood constraint~\cite{wang2018learning,kim2020mule} to the output query and subtitle embeddings. 
It encourages sentences of the same meaning in different languages to lie close to each other in the embedding space.
Compared to separately trained monolingual models, \ModelName~substantially reduces the total model size while improving retrieval performance (over monolingual models) as we show in Section~\ref{sec:experiments}. 
Detailed dataset analyses and model ablations are provided.

\section{Dataset}\label{sec:dataset}
The TVR~\cite{lei2020tvr} dataset contains 108,965 high-quality English queries from 21,793 videos from 6 long-running TV shows (provided by TVQA~\cite{Lei2018TVQALC}). The videos are associated with English dialogues in the form of subtitle text. \DsetName~extends this dataset with translated dialogues and queries in Chinese to support multilingual multimodal research.

\subsection{Data Collection}

\paragraph{Dialogue Subtitles.}
We crawl fan translated Chinese subtitles from subtitle sites.\footnote{\url{https://subhd.tv}, \url{http://zimuku.la}} 
All subtitles are manually checked by the authors to ensure they are of good quality and are aligned with the videos.
The original English subtitles come with speaker names from transcripts that we map to the Chinese subtitles, to ensure that the Chinese subtitles have the same amount of information as the English version.

\paragraph{Query.}
To obtain Chinese queries, we hire human translators from Amazon Mechanical Turk (AMT).
Each AMT worker is asked to write a Chinese translation of a given English query.
Languages are ambiguous, hence we also present the original videos to the workers at the time of translation to help clarify query meaning via spatio-temporal visual grounding. The Chinese translations are required to have the exact same meaning as the original English queries and the translation should be made based on the aligned video content.
To facilitate the translation process, we provide machine translated Chinese queries from Google Cloud Translation\footnote{\url{https://cloud.google.com/translate}} as references, similar to~\cite{wang2019vatex}. 
To find qualified bilingual workers in AMT, we created a qualification test with 5 multiple-choice questions designed to evaluate workers' Chinese language proficiency and their ability to perform our translation task.
We only allow workers that correctly answer all 5 questions to participate our annotation task.
In total, 99 workers finished the test and 44 passed, earning our qualification.
To further ensure data quality, we also manually inspect the submitted results during the annotation process and disqualify workers with poor annotations.
We pay workers \$0.24 every three sentences, this results in an average hourly pay of \$8.70. 
The whole annotation process took about 3 months and cost approximately \$12,000.00.

\begin{table}[!t]
\centering
\small
\setlength{\tabcolsep}{2pt}
\renewcommand{\arraystretch}{1.3}
\scalebox{0.87}{
\begin{CJK*}{UTF8}{gbsn}
\begin{tabular}{ll}
\toprule
QType (\%) & \multicolumn{1}{c}{Query Examples (in English and Chinese)} \\
\midrule
video-only & Howard places his plate onto the coffee table. \\
(74.2)  & 霍华德将盘子放在咖啡桌子上。\\
\midrule
sub-only & Alexis and Castle talk about the timeline of the murder. \\
 (9.1) & 
亚历克西斯和卡塞尔谈论谋杀的时间顺序。
 \\
\midrule
video+sub & Joey waives his hand when he asks for his food.  \\
(16.6) & 
乔伊催餐时摆了摆手。\\
\bottomrule
\end{tabular}
\end{CJK*} 
}
\caption{
\DsetName~English and Chinese query examples in different query types. 
The percentage of the queries in each query type is shown in brackets.
}
\label{tab:qtype_examples}
\end{table}

\subsection{Data Analysis}
In Table~\ref{tab:en_zh_data_comparison}, we compare the average sentence lengths and the number of unique words under different part-of-speech (POS) tags, between the two languages, English and Chinese, and between query and subtitle text.
For both languages, dialogue subtitles are linguistically more diverse than queries, i.e., they have more unique words in all categories. 
This is potentially because the language used in subtitles are unconstrained human dialogues while the queries are collected as declarative sentences referring to specific moments in videos~\cite{lei2020tvr}. 
Comparing the two languages, the Chinese data is typically more diverse than the English data.\footnote{
\begin{CJK*}{UTF8}{gbsn}
The differences might be due to the different morphemes in the languages.
E.g., the Chinese word 长发\; (`long hair') is labeled as a single noun, but as an adjective (`long') and a noun (`hair') in English~\cite{wang2019vatex}.
\end{CJK*}
}
In Table~\ref{tab:qtype_examples}, we show English and their translated Chinese query examples in Table~\ref{tab:qtype_examples}, by query type.
In the appendix, we compare \DsetName~with existing video and language datasets.

\begin{table}[]
\centering
\small
\scalebox{0.86}{
\begin{tabular}{lrrrrrr}
\toprule
\multirow{2}{*}{ Data } & \multicolumn{1}{c}{Avg} & \multicolumn{5}{c}{ \#unique words by POS tags } \\ \cmidrule(l){3-7}
& \multicolumn{1}{c}{Len} & \multicolumn{1}{c}{all} & \multicolumn{1}{c}{verb}  & \multicolumn{1}{c}{noun} & \multicolumn{1}{c}{adj.} & \multicolumn{1}{c}{adv.} \\
\midrule
\multicolumn{2}{l}{\textbf{English}} & & & &  &  \\
Q & 13.45 & 15,201 & 3,015 & 7,143 & 2,290 & 763 \\
Sub & 10.78 & 49,325 & 6,441 & 19,223 & 7,504 & 1,740 \\
Q+Sub & 11.27 & 52,545 & 7,151 & 20,689 & 8,021 & 1,976 \\
\midrule
\multicolumn{2}{l}{\textbf{Chinese}} & & & &  &  \\
Q & 12.55 & 34,752 & 12,773 & 18,706 & 1,415 & 1,669 \\
Sub & 9.04 & 101,018 & 36,810 & 53736 & 4,958 & 5,568 \\
Q+Sub & 9.67 & 117,448 & 42,284 & 62,611 & 5,505 & 6,185 \\
\bottomrule
\end{tabular}
}
\vspace{-3pt}
\caption{Comparison of English and Chinese data in~\DsetName. We show average sentence length, and number of unique tokens by POS tags, for Query (\textit{Q}) and or Subtitle (\textit{Sub}).}
\label{tab:en_zh_data_comparison}
\vspace{-8pt}
\end{table}

\section{Method}\label{method}
Our multilingual moment retrieval model \ModelName~is built on top of the Cross-model Moment Localization (XML)~\cite{lei2020tvr} model, which performs efficient video-level retrieval at its shallow layers and accurate moment-level retrieval at its deep layers. 
To adapt the monolingual XML model into the multilingual setting in \DsetName~and improve its efficiency and effectiveness, we apply encoder parameter sharing and neighborhood constraints~\cite{wang2018learning,kim2020mule} which encourages the model to better utilize multilingual data to improve monolingual task performance while maintaining smaller model size.

\paragraph{Query and Context Representations.}
We represent videos using ResNet-152~\cite{he2016deep} and I3D~\cite{carreira2017quo} features extracted every 1.5 seconds. 
We extract language features using pre-trained, then finetuned (on our queries and subtitles) RoBERTa-base~\cite{liu2019roberta}, for English~\cite{liu2019roberta} and Chinese~\cite{cui2020revisiting}, respectively.
For queries, we use token-level features. 
For subtitles, we max-pool the token-level features every 1.5 seconds to align with the video features. 
We then project the extracted features into a low-dimensional space via a linear layer, and add learned positional encoding~\cite{devlin2018bert} after the projection. 
We denote the resulting video features as $E^v \in \mathbb{R}^{l \times d}$, subtitle features as $E^s_{en} \in \mathbb{R}^{l \times d}, E^s_{zh} \in \mathbb{R}^{l \times d}$, and query features as $E^{q}_{en} \in \mathbb{R}^{l_q \times d}, E^{q}_{zh} \in \mathbb{R}^{l_q \times d}$. $l$ is video length, $l_q$ is query length, and $d$ is hidden size. The subscripts $en$ and $zh$ denote English and Chinese text features, respectively. 

\begin{figure}[!t]
\begin{center}
  \includegraphics[width=0.9\columnwidth]{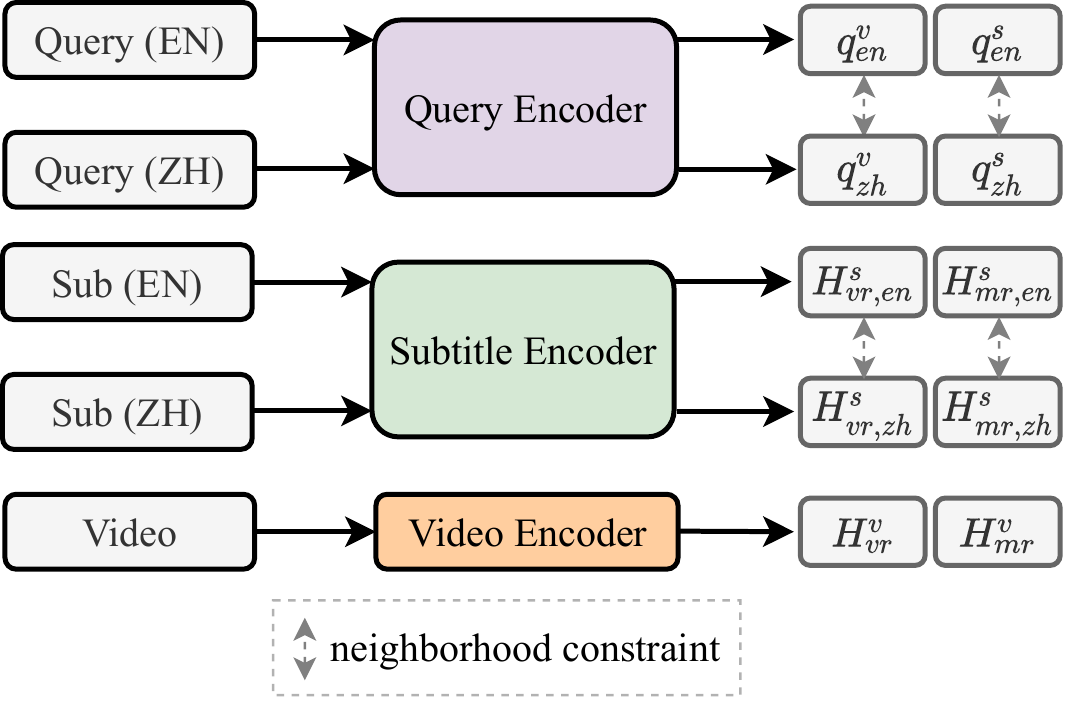}
  \vspace{-3pt}
  \caption{
  Illustration of \ModelName's encoding process.
  Compared to monolingual models, \ModelName~learns from the two languages simultaneously, and allows them to benefit each other via \textit{encoder parameter sharing} and \textit{neighborhood constraints}. 
  We show the detailed encoding process of the model in the appendix (Figure~\ref{fig:mxml_overview}).
  }
  \label{fig:tvrm_encoding}
  \vspace{-8pt}
  \end{center}
\end{figure}

\paragraph{Encoders and Parameter Sharing.}
We follow~\citet{lei2020tvr} to use \textit{Self-Encoder} as our main component for query and context encoding. 
A Self-Encoder consists of a self-attention~\cite{vaswani2017attention} layer, a linear layer, and a residual~\cite{he2016deep} connection followed by layer normalization~\cite{ba2016layer}.
We use a Self-Encoder followed by a modular attention~\cite{lei2020tvr} to encode each query into two modularized query vectors $\boldsymbol{q}^{v}_{lang}, \boldsymbol{q}^{s}_{lang} \in \mathbb{R}^{d}$ ($lang\in \{en, zh\}$) for video and subtitle retrieval, respectively.
For videos, we apply two Self-Encoders instead of a Self-Encoder and a Cross-Encoder as in XML, because we found this modification simplifies the implementation while maintains the performance.
We use the outputs from the first and the second Self-Encoder $H^v_{vr,lang}, H^v_{mr,lang} \in \mathbb{R}^{l \times d}$ for video retrieval and moment retrieval. 
Similarly, we have $H^s_{vr,lang},H^s_{mr,lang} \in \mathbb{R}^{l \times d}$ for subtitles.
All the Self-Encoders are shared across languages, e.g., we use the same Self-Encoder to encode both English and Chinese queries, as illustrated in Figure~\ref{fig:tvrm_encoding}. 
This parameter sharing strategy greatly reduces the model size while maintaining or even improving model performance, as we show in Section~\ref{sec:experiments}.

\paragraph{Language Neighborhood Constraint.}
To facilitate stronger multilingual learning, we add neighborhood constraints~\cite{wang2018learning,kim2020mule,burns2020learning} to the model. 
This encourages sentences that express the same or similar meanings to lie close to each other in the embedding space, via a triplet loss.
Given paired sentence embeddings $\boldsymbol{e}_{en}^{i} \in \mathbb{R}^d$ and $\boldsymbol{e}_{zh}^{i} \in \mathbb{R}^d$, we sample negative sentence embeddings $\boldsymbol{e}_{en}^{j} \in \mathbb{R}^d$ and $\boldsymbol{e}_{zh}^{k} \in \mathbb{R}^d$ from the same mini-batch, where $i \neq j, i \neq k$.
We use cosine similarity function $\mathcal{S}$ to measure the similarity between embeddings.
Our language neighborhood constraint can be formulated as:
\scalebox{0.9}{\parbox{1.1\columnwidth}{%
\begin{align}\label{eq:triplet}
    \mathcal{L}_{nc} &{=} \frac{1}{n}\sum_i[\mathrm{max}(0, \mathcal{S}(\boldsymbol{e}_{en}^{i}, \boldsymbol{e}_{zh}^{k}) - \mathcal{S}(\boldsymbol{e}_{en}^{i}, \boldsymbol{e}_{zh}^{i}) + \Delta) \nonumber \\&+ \mathrm{max}(0, \mathcal{S}(\boldsymbol{e}_{en}^{j}, \boldsymbol{e}_{zh}^{i}) - \mathcal{S}(\boldsymbol{e}_{en}^{i}, \boldsymbol{e}_{zh}^{i}) + \Delta)],
\end{align}
}}

\noindent
where $\Delta{=}0.2$ is the margin.
We apply this constraint on both query and subtitle embeddings, across the two languages, as illustrated in Figure~\ref{fig:tvrm_encoding}. For queries, we directly apply it on the query vectors $\boldsymbol{q}^{v}_{lang}, \boldsymbol{q}^{s}_{lang}$. For the subtitle embeddings, we apply it on the embeddings $H^{s}_{vr,lang}, H^{s}_{mr,lang}$, after max-pooling them in the temporal dimension.

\paragraph{Training and Inference.}
During training, we optimize video retrieval scores with a triplet loss, and moment scores with a cross-entropy loss. 
At inference, these two scores are aggregated together as the final score for video corpus moment retrieval.
See appendix for details.

\section{Experiments and Results}\label{sec:experiments}
We evaluate our proposed \ModelName~model on the newly collected \DsetName~dataset, and compare it with several existing monolingual baselines. We also provide ablation studies evaluating our model design and the importance of each input modality (videos and subtitles).

\noindent\textbf{Data Splits and Evaluation Metrics.}
We follow TVR~\cite{lei2020tvr} to split the data into 80\% \textit{train}, 10\% \textit{val}, 5\% \textit{test-public} and 5\% \textit{test-private}. We report average recall (R@1) on the Video Corpus Moment Retrieval (VCMR) task. A predicted moment is correct if it has high Intersection-over-Union (IoU) with the ground-truth.

\begin{table}[]
\centering
\small
\setlength{\tabcolsep}{1.5pt}
\renewcommand{\arraystretch}{1.05}
\scalebox{1.0}{
\begin{tabular}{lccccc}
\toprule
\multirow{2}{*}{ Method } & \multirow{2}{*}{ \#param } & \multicolumn{2}{c}{ English R@1 } & \multicolumn{2}{c}{ Chinese R@1 } \\
\cmidrule(l){3-4} \cmidrule(l){5-6}
& & IoU=0.5 & IoU=0.7 & IoU=0.5 & IoU=0.7 \\
\midrule
Chance & - & 0.00 & 0.00 & 0.00 & 0.00 \\
\multicolumn{3}{l}{\textbf{Proposal based}} & & & \\
MCN & 6.4M & 0.02 & 0.00 & 0.13 & 0.02 \\
CAL & 6.4M & 0.09 & 0.04 & 0.11 & 0.04 \\
\multicolumn{3}{l}{\textbf{Retrieval + Re-ranking}} & & & \\
MEE+MCN & 10.4M & 0.92 & 0.42 & 1.43 & 0.64 \\
MEE+CAL & 10.4M & 0.97 & 0.39 & 1.51 & 0.62 \\
MEE+ExCL & 10.0M & 0.92 & 0.33 & 1.43 & 0.72 \\
XML & 6.4M & 7.25 & 3.25 & 5.91 & 2.57 \\
\midrule
\bf \ModelName & \bf 4.5M & \bf 8.30 & \bf 3.82 & \bf 6.76 & \bf 3.20 \\
\bottomrule
\end{tabular}
}
\vspace{-3pt}
\caption{
Baseline comparison on \DsetName~\textit{test-public} split. \ModelName~achieves better retrieval performance on both languages while using fewer parameters.}
\label{tab:vcmr_baseline_comparison}
\vspace{-3pt}
\end{table}

\noindent\textbf{Baseline Comparison.}
In Table~\ref{tab:vcmr_baseline_comparison}, we compare \ModelName~with multiple baseline approaches. 
Given a natural language query, the goal of video corpus moment retrieval is to retrieve relevant moments from a large video corpus.
The methods for this task can be grouped into two categories, ($i$) proposal based approaches (MCN~\cite{anne2017localizing} and CAL~\cite{escorcia2019temporal}), where they perform video retrieval on the pre-segmented moments from the videos; ($ii$) retrieval+re-ranking methods (MEE~\cite{miech2018learning}+MCN, MEE+CAL, MEE+ExCL~\cite{ghosh2019excl} and XML~\cite{lei2020tvr}), where one approach is first used to retrieve a set of videos, then another approach is used to re-rank the moments inside these retrieved videos to get the final moments.
Our method \ModelName~also belongs to the retrieval+re-ranking category.
Across all metrics and both languages, we notice retrieval+re-ranking approaches achieve better performance than proposal based approaches, indicating that retrieval+re-ranking is potentially better suited for the VCMR task.
Meanwhile, our \ModelName~outperforms the strong baseline XML significantly\footnote{Statistically significant with $p{<}0.01$. We use bootstrap test~\cite{efron1994introduction,noreen1989computer}.} while using few parameters.
XML is a monolingual model, where a separate model is trained for each language. 
In contrast, \ModelName~is multilingual, trained on both languages simultaneously, with parameter sharing and language neighborhood constraints to encourage multilingual learning. 
\ModelName~prediction examples are provided in the appendix.

\paragraph{Ablations on Model Design.} 
In Table~\ref{tab:ablation_model_design}, we present our ablation study on~\ModelName. 
We use `\textit{Baseline}' to denote the~\ModelName{} model without parameter sharing and neighborhood constraint. 
Sharing encoder parameter across languages greatly reduces \#parameters while maintaining (Chinese) or even improving (English) model performance.
Adding neighborhood constraint does not introduce any new parameters but brings a notable ($p{<}0.06$) performance gain to both languages.
We hypothesize that this is because the learned information in the embeddings of the two languages are complementary (though the sentences in the two languages express the same meaning, their language encoders~\cite{liu2019roberta,cui2020revisiting}) are pre-trained differently, which may lead to different meanings at the embedding level.
In Table~\ref{tab:ablation_query_type_comparison}, we show a detailed comparison between~\ModelName{} and its baseline version, by query types. 
Overall, we notice the \ModelName~perform similarly with \textit{Baseline} in `\textit{video}' queries, but shows a significant performance gain in `\textit{subtitle}' queries, suggesting the parameter sharing and neighborhood constraint are more useful for queries that need more language understanding.

\begin{table}[]
\centering
\small
\setlength{\tabcolsep}{1pt}
\renewcommand{\arraystretch}{1.05}
\scalebox{0.96}{
\begin{tabular}{lccccc}
\toprule
\multirow{2}{*}{ Method } & \multirow{2}{*}{ \#param } & \multicolumn{2}{c}{ English R@1 } & \multicolumn{2}{c}{ Chinese R@1 } \\
\cmidrule(l){3-4} \cmidrule(l){5-6}
& & IoU=0.5 & IoU=0.7 & IoU=0.5 & IoU=0.7 \\
\midrule
Baseline & 6.4M & 5.77 & 2.63 & 4.7 & 2.38 \\
\;+ Share Enc. & \bf 4.5M & 6.09 & 2.85 & 4.72 & 2.25 \\
\;\;\;+ NC (\ModelName) & \bf 4.5M & \bf 6.22 & \bf 2.96 & \bf 5.17 & \bf 2.41 \\
\bottomrule
\end{tabular}
}
\vspace{-3pt}
\caption{
\ModelName~ablation study on \DsetName~\textit{val} split. \textit{Share Enc.} {=} encoder parameter sharing, \textit{NC} {=} Neighborhood Constraint. 
Each row adds an extra component to the row above it.
}
\label{tab:ablation_model_design}
\vspace{-6pt}
\end{table}

\begin{table}[!t]
\centering
\small
\setlength{\tabcolsep}{2.5pt}
\renewcommand{\arraystretch}{1.05}
\scalebox{1.0}{
\begin{tabular}{lcccc}
\toprule
Model Type & \multicolumn{2}{c}{English R@1} & \multicolumn{2}{c}{Chinese R@1} \\  \cmidrule(l){2-3} \cmidrule(l){4-5}
& IoU=0.5 & IoU=0.7 & IoU=0.5 & IoU=0.7 \\
\midrule
\multicolumn{2}{l}{\textbf{Query type: video}} & & \\
Baseline & 5.46 & 2.53 & 4.78 & 2.47 \\
mXML & 5.77 & 2.67 & 5.14 & 2.32 \\
\midrule
\multicolumn{2}{l}{\textbf{Query type: subtitle}} & & \\
Baseline & 4.15 & 1.97 & 3.11 & 1.14 \\
mXML & 6.12 & 3.32 & 4.05 & 1.87 \\
\midrule
\multicolumn{2}{l}{\textbf{Query type: video+subtitle}} & & \\
baseline & 8.02 & 3.38 & 5.18 & 2.62 \\
mXML & 8.29 & 4.09 & 5.89 & 3.11 \\
\bottomrule
\end{tabular}
}
\caption{Comparison of \ModelName~and the baseline on \DsetName~\textit{val} set, with breakdown on query types. Both models are trained with video and subtitle as inputs.}
\label{tab:ablation_query_type_comparison}
\end{table}

\paragraph{Ablations on Input Modalities.}
In Table~\ref{tab:ablation_input_modalities_query_type}, we compare \ModelName~variants with different context inputs, i.e., video or subtitle or both.
We report their performance under the three annotated query types, \textit{video}, \textit{sub} and \textit{video+sub}.
Overall, the model with both video and subtitle as inputs perform the best.
The video model performs much better on the \textit{video} queries than on the \textit{sub} queries, while the subtitle model achieves higher scores on the \textit{sub} queries than the \textit{video} queries.

In the appendix, we also present results on `\textit{generalization to unseen TV shows}' setup. 

\begin{table}[!t]
\centering
\small
\setlength{\tabcolsep}{2pt}
\renewcommand{\arraystretch}{1.05}
\scalebox{1.0}{
\begin{tabular}{lcccc}
\toprule
\multirow{2}{*}{ QType (percentage) } & \multicolumn{2}{c}{ English R@1 } & \multicolumn{2}{c}{ Chinese R@1 } \\ \cmidrule(l){2-3} \cmidrule(l){4-5}
& IoU=0.5 & IoU=0.7 & IoU=0.5 & IoU=0.7 \\
\midrule
\multicolumn{1}{l}{\textbf{Model input: video}} & & & & \\
video (74.32\%) & 4.12 & 1.89 & 3.73 & 1.86 \\
sub (8.85\%) & 1.97 & 1.24 & 1.35 & 1.04 \\
video+sub (16.83\%)  & 2.67 & 1.2 & 2.45 & 1.15 \\
\midrule
\multicolumn{3}{l}{\textbf{Model input: subtitle}} & & \\
video & 1.35 & 0.62 & 1.11 & 0.51 \\
sub & 6.33 & 2.9 & 4.15 & 1.97 \\
video+sub & 6.22 & 2.62 & 4.2 & 2.13 \\
\midrule
\multicolumn{3}{l}{\textbf{Model input: video+subtitle}} & & \\
video & 5.77 & 2.67 & 5.14 & 2.32 \\
sub & 6.12 & 3.32 & 4.05 & 1.87 \\
video+sub & 8.29 & 4.09 & 5.89 & 3.11 \\
\bottomrule
\end{tabular}
}
\caption{\ModelName~performance breakdown on \DsetName~\textit{val} set by query types, with different inputs. }
\label{tab:ablation_input_modalities_query_type}
\end{table}

\section{Conclusion}\label{conclusion}
In this work, we collect \DsetName, a new large-scale, multilingual moment retrieval dataset. 
It contains 218K queries in English and in Chinese from 21.8K video clips from 6 TV shows. 
We also propose a multilingual moment retrieval model \ModelName~as a strong baseline for the \DsetName~dataset. 
We show in experiments that \ModelName~outperforms monolingual models while using fewer parameters.

\section*{Acknowledgements}
We thank the reviewers for their helpful feedback. This research is supported by NSF Award \#1562098, DARPA KAIROS Grant \#FA8750-19-2-1004, and ARO-YIP Award \#W911NF-18-1-0336. The views contained in this article are those of the authors and not of the funding agency.

\bibliographystyle{acl_natbib}
\bibliography{anthology,acl2021}

\appendix

\section{Appendix}

\paragraph{Data Analysis.}
In Table~\ref{tab:dataset_comparison} we show a comparison of \DsetName~with existing moment retrieval datasets and related video and language datasets. 
Compared to other moment retrieval datasets, \DsetName~is significantly larger in scale, and comes with query type annotations that allows in-depth analyses for the models trained on it.
Besides, it is also the only moment retrieval dataset with multilingual annotations, which is vital in studying the moment retrieval problem under the multilingual context. 
Compared to the existing multilingual video and language datasets, \DsetName~is unique as it has a more diverse set of context and annotations, i.e., dialogue, query type, and timestamps.

\begin{figure*}[!t]
  \includegraphics[width=\linewidth]{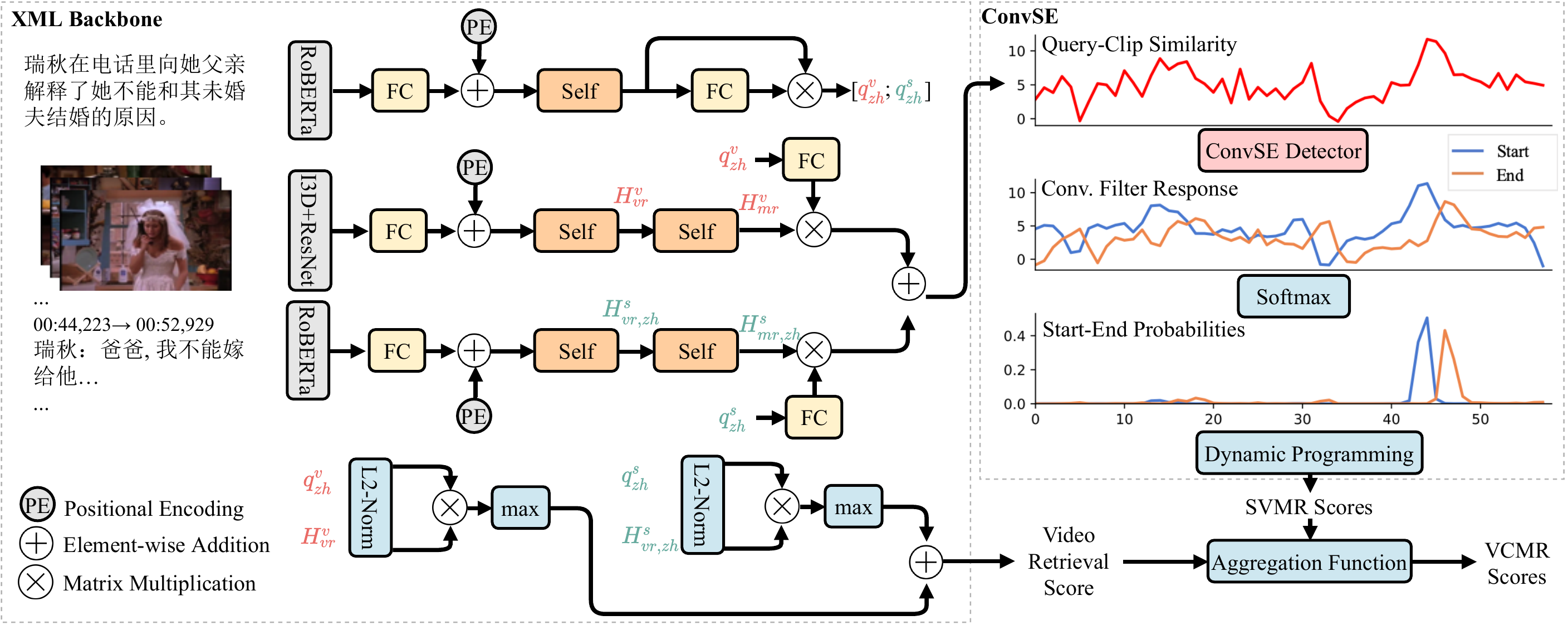}
  \caption{
  \ModelName~overview. For brevity, we only show the modeling process for a single language (Chinese). The cross-language modifications, i.e., parameter sharing and neighborhood constraint are illustrated in Figure~\ref{fig:tvrm_encoding}. This figure is edited from the Figure 4 in~\citep{lei2020tvr}. 
  }
  \label{fig:mxml_overview}
\end{figure*}

\paragraph{Training and Inference Details.}
In Figure~\ref{fig:mxml_overview} we show an overview of the \ModelName~model.
We compute video retrieval score as:
\begin{align}
    s^{vr} = \frac{1}{2}\sum_{m \in \{v, s\}} \mathrm{max}(\frac{H^{m}_{vr}}{\left\Vert H^{m}_{vr}\right\Vert} \frac{\boldsymbol{q}^{m}}{\left\Vert \boldsymbol{q}^{m}\right\Vert}).
\end{align}
The subscript $lang \in \{en, zh\}$ is omitted for simplicity.
It is optimized using a triplet loss similar to main text Equation (1).
For moment retrieval, we first compute the query-clip similarity scores $S^{q,c} \in \mathbb{R}^{l}$ as:
\begin{align}
    S^{q,c} = \frac{1}{2}(H^{s}_{mr}\boldsymbol{q}^{s} + H^{v}_{mr}\boldsymbol{q}^{v}).
\end{align}
Next, we apply Convolutional Start-End Detector (ConvSE module)~\cite{lei2020tvr} to obtain start, end probabilities $P_{st}, P_{ed} \in \mathbb{R}^{l}$. These scores are optimized using a cross-entropy loss. The single video moment retrieval score for moment $[t_{st}, t_{ed}]$ is computed as:
\begin{align}
    s^{mr}(t_{st}, t_{ed}) = P_{st}(t_{st}) P_{ed}(t_{ed}), \, t_{st} \leq t_{ed}.
\end{align}

\noindent
Given a query $q_i$, the retrieval score for moment [$t_{st}$:$t_{ed}$] in video $v_j$ is computed following the aggregation function as in~\cite{lei2020tvr}:
\begin{align}
    s^{vcmr}&(v_j,t_{st}, t_{ed}|q_i) = \nonumber \\ &s^{mr}(t_{st}, t_{ed})\mathrm{exp}(\alpha s^{vr}(v_j|q_i)),
\end{align}

\noindent
where $\alpha{=}20$ is used to assign higher weight to the video retrieval scores.
The overall loss is a simple summation of video and moment retrieval loss across the two languages, and the language neighborhood constraint loss.

\paragraph{Implementation Details.}
\ModelName~is implemented in PyTorch~\cite{paszke2017automatic}.
We use Adam~\cite{kingma2014adam} with initial learning rate 1e-4, $\beta_1{=}0.9$, $\beta_2{=}0.999$, L2 weight decay 0.01, learning rate warm-up over the first 5 epochs. 
We train \ModelName~for at most 100 epochs at batch size 128, with early stop based on the sum of R@1 (IoU=0.7) scores for English and Chinese.
The experiments are conducted on a NVIDIA RTX 2080Ti GPU. 
Each run takes around 7 hours.

\begin{table}[!t]
\centering
\small
\setlength{\tabcolsep}{3.5pt}
\renewcommand{\arraystretch}{1.05}
\scalebox{1.0}{
\begin{tabular}{lcccc}
\toprule
& \multicolumn{2}{c}{English R@1} & \multicolumn{2}{c}{Chinese R@1} \\  \cmidrule(l){2-3} \cmidrule(l){4-5}
Setting & IoU=0.5 & IoU=0.7 & IoU=0.5 & IoU=0.7 \\
\midrule
unseen & 1.68 & 0.79 & 1 & 0.54 \\
seen & 4.82 & 2.79 & 4.18 & 2.32 \\
\bottomrule
\end{tabular}
}
\caption{\ModelName~performance on the \DsetName~val split \textit{Friends} examples, in both \textit{unseen} and \textit{seen} settings. 
}
\label{tab:ablation_unseen}
\end{table}

\begin{table*}[ht]
\centering
\small
\setlength{\tabcolsep}{5pt}
\scalebox{0.96}{
\begin{tabular}{lcccccc}
\toprule
Dataset & Domain & \#Q/\#videos & Multilingual & Dialogue & QType & Timestamp \\
\midrule
\bf QA datasets with temporal annotation &  &  &  &  &  &  \\
TVQA~\cite{Lei2018TVQALC} & TV show & 152.5K/21.8K & - & \checkmark & - & \checkmark \\
How2QA~\cite{li2020hero} & Instructional & 44K/22K & - & \checkmark & - & \checkmark \\
\bf Multilingual video description datasets &  &  &  &  &  &  \\
MSVD~\cite{chen2011collecting} & Open & 70K/2K & \checkmark & - & - & - \\
VATEX~\cite{wang2019vatex} & Activity & 826K/41.3K & \checkmark & - & - & - \\
\bf Moment retrieval datasets &  &  &  &  &  &  \\
TACoS~\cite{regneri2013grounding} & Cooking & 16.2K/0.1K & - & - & - & \checkmark \\
DiDeMo~\cite{anne2017localizing} & Flickr & 41.2K/10.6K & - & - & - & \checkmark \\
ActivityNet Captions~\cite{Krishna2017DenseCaptioningEI} & Activity & 72K/15K & - & - & - & \checkmark \\
CharadesSTA~\cite{gao2017tall} & Activity & 16.1K/6.7K & - & - & - & \checkmark \\
How2R~\cite{li2020hero} & Instructional & 51K/24K & - & \checkmark & - & \checkmark \\
TVR~\cite{lei2020tvr} & TV show & 109K/21.8K & - & \checkmark & \checkmark & \checkmark \\
\midrule
\DsetName & TV show & 218K/21.8K & \checkmark & \checkmark & \checkmark & \checkmark \\ 
\bottomrule
\end{tabular}
}
\caption{
Comparison of~\DsetName~with related video and language datasets.   
}
\label{tab:dataset_comparison}
\end{table*}

\paragraph{Generalization to Unseen TV shows.} 
To investigate whether the learned model can be transferred to other TV shows, we conduct an experiment by using the TV show `\textit{Friends}' as an `\textit{unseen}' TV show for testing, and train the model on all the other 5 TV shows. 
For comparison, we also include a model trained on `\textit{seen}' setting, where we use all the 6 TV shows including \textit{Friends} for training. 
To ensure the models on these two settings are trained on the same number of examples, we downsample the examples in the \textit{seen} setting to match the \textit{unseen} setting.
The results are shown in Table~\ref{tab:ablation_unseen}.
We notice our \ModelName~achieves a reasonable performance even though it does see a single example from the TV show \textit{Friends}.
Meanwhile, the gap between \textit{unseen} and \textit{seen} settings are still large, we encourage future work to further explore this direction.

\begin{figure*}[!t]
  \includegraphics[width=\linewidth]{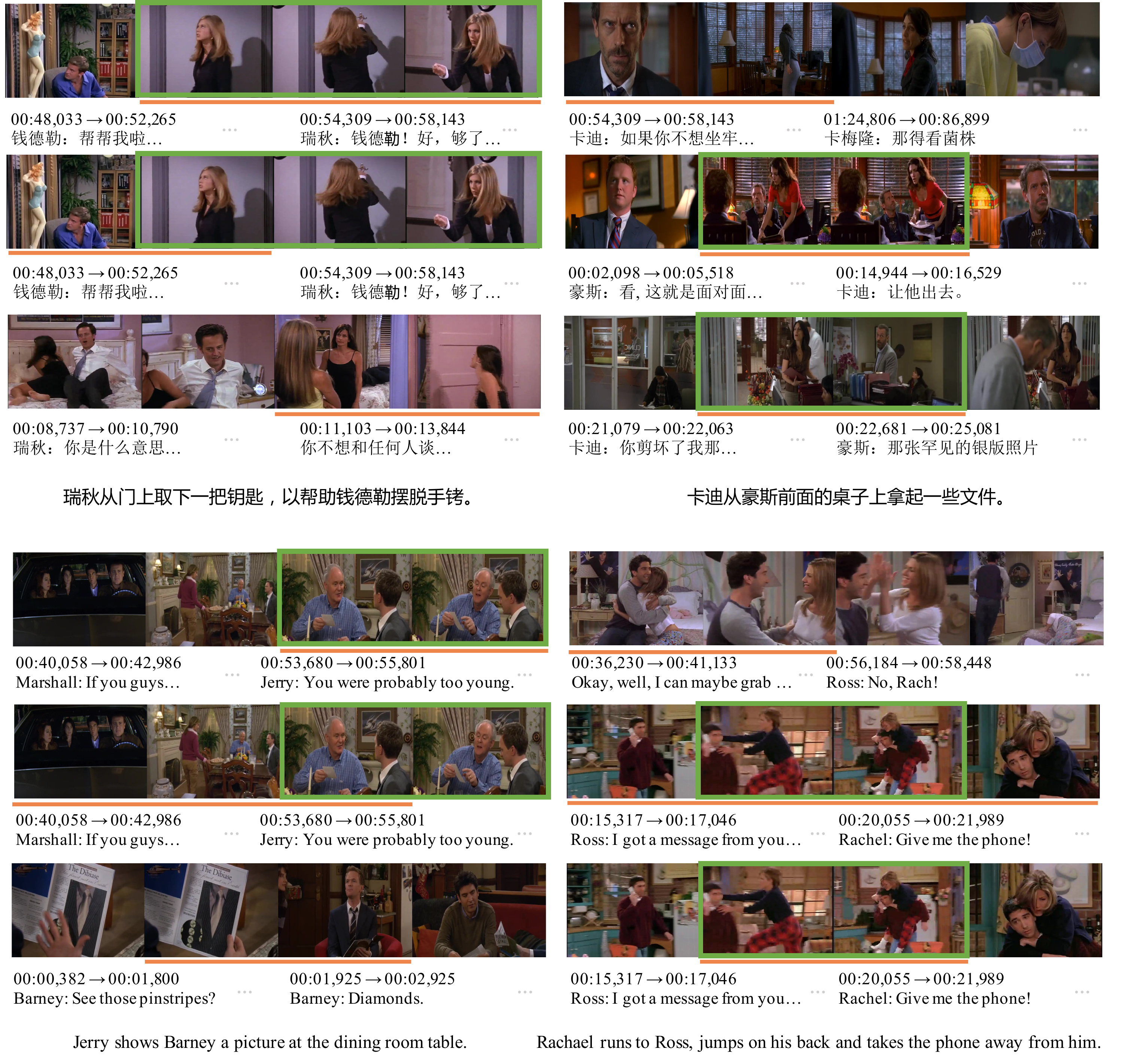}
  \caption{
  Qualitative examples of \ModelName. \textit{Top:} examples in Chinese. \textit{Bottom:} examples in English. \textit{Left:} correct predictions. \textit{Right:} incorrect predictions.
  We show top-3 retrieved moments for each query. \textcolor{salmon}{salmon bar} shows the predictions, \textcolor{ForestGreen}{green box} indicates the ground truth.
  }
  \label{fig:pred_examples}
\end{figure*}

\paragraph{Prediction Examples}
We show \ModelName~prediction examples in Figure~\ref{fig:pred_examples}. 
We show both Chinese (\textit{top}) and English (\textit{bottom}) prediction examples, and correct (\textit{left}) and incorrect (\textit{right}) examples.

\end{document}